%% file: main.tex
\documentclass[10pt,twocolumn,letterpaper]{article}

\usepackage{cvpr}              

\usepackage{graphicx}
\usepackage{amsmath}
\usepackage{amssymb}
\usepackage{booktabs}

\usepackage{multirow}
\usepackage{floatrow}

\usepackage{tabularx}

\DeclareMathOperator*{\argmin}{arg\,min}

\usepackage{hyperref}
\hypersetup{pagebackref,breaklinks,colorlinks}

\usepackage[capitalize]{cleveref}
\crefname{section}{Sec.}{Secs.}
\Crefname{section}{Section}{Sections}
\Crefname{table}{Table}{Tables}
\crefname{table}{Tab.}{Tabs.}

\def\cvprPaperID{6872} 
\def\confName{CVPR}
\def\confYear{2023}

\begin{document}

\title{Learning a Depth Covariance Function}

\author{Eric Dexheimer and Andrew J. Davison\\
Dyson Robotics Lab, Imperial College London\\
{\tt\small \{e.dexheimer21, a.davison\}@imperial.ac.uk }
}
\maketitle

\input{sections/abstract.tex}

\input{sections/introduction.tex}
\input{sections/related_work.tex}
\input{sections/learning_depth_cov.tex}
\input{sections/leveraging_depth_cov.tex}
\input{sections/applications.tex}
\input{sections/conclusion.tex}

\section{Acknowledgements}
Research presented in this paper has been supported by Dyson Technology Ltd.  We would like to thank Tristan Laidlow, Hidenobu Matsuki, Riku Murai, and members of the Dyson Robotics Lab for insightful discussions.

{\small
\bibliographystyle{ieee_fullname}
\bibliography{bibmap}
}

\newpage
\appendix
\input{sections/supp.tex}

\end{document}

%% file: sections/abstract.tex
\begin{abstract}
   We propose learning a depth covariance function with applications to geometric vision tasks. Given RGB images as input, the covariance function can be flexibly used to define priors over depth functions, predictive distributions given observations, and methods for active point selection. We leverage these techniques for a selection of downstream tasks: depth completion, bundle adjustment, and monocular dense visual odometry.
\end{abstract}

%% file: sections/introduction.tex
\section{Introduction}
\label{sec:intro}

Inferring the 3D structure of the world from 2D images is an essential computer vision task. In recent years, there has been significant interest in combining principled multiple view geometry with data-driven priors. Learning-based methods that predict geometry provide a prior directly over the latent variables, which avoids the ill-posed configurations of traditional methods.  However, direct, overconfident priors may prevent realization of the true 3D structure when multi-view geometry is well-defined.  For example, data-driven methods have shown tremendous promise in monocular depth estimation, but often lack consistency when fusing information into 3D space.

Thus far, designing a unified framework that combines the best of learning and optimization methods has proven challenging.  Recent data-driven methods have attempted to relax rigid geometric constraints by also predicting low-dimensional subspaces or per-pixel uncertainties.  Fixing capacity during training is often wasteful or inflexible at test time, while per-pixel residual distributions typically explain away the limitations of the model instead of the relationship between errors.  In reality, the 3D world is anywhere from simple to complex, and an ideal system should explicitly adapt its capacity and correlation based on the scene.

\begin{figure}[t]
	\centering
	\includegraphics[width=\columnwidth]{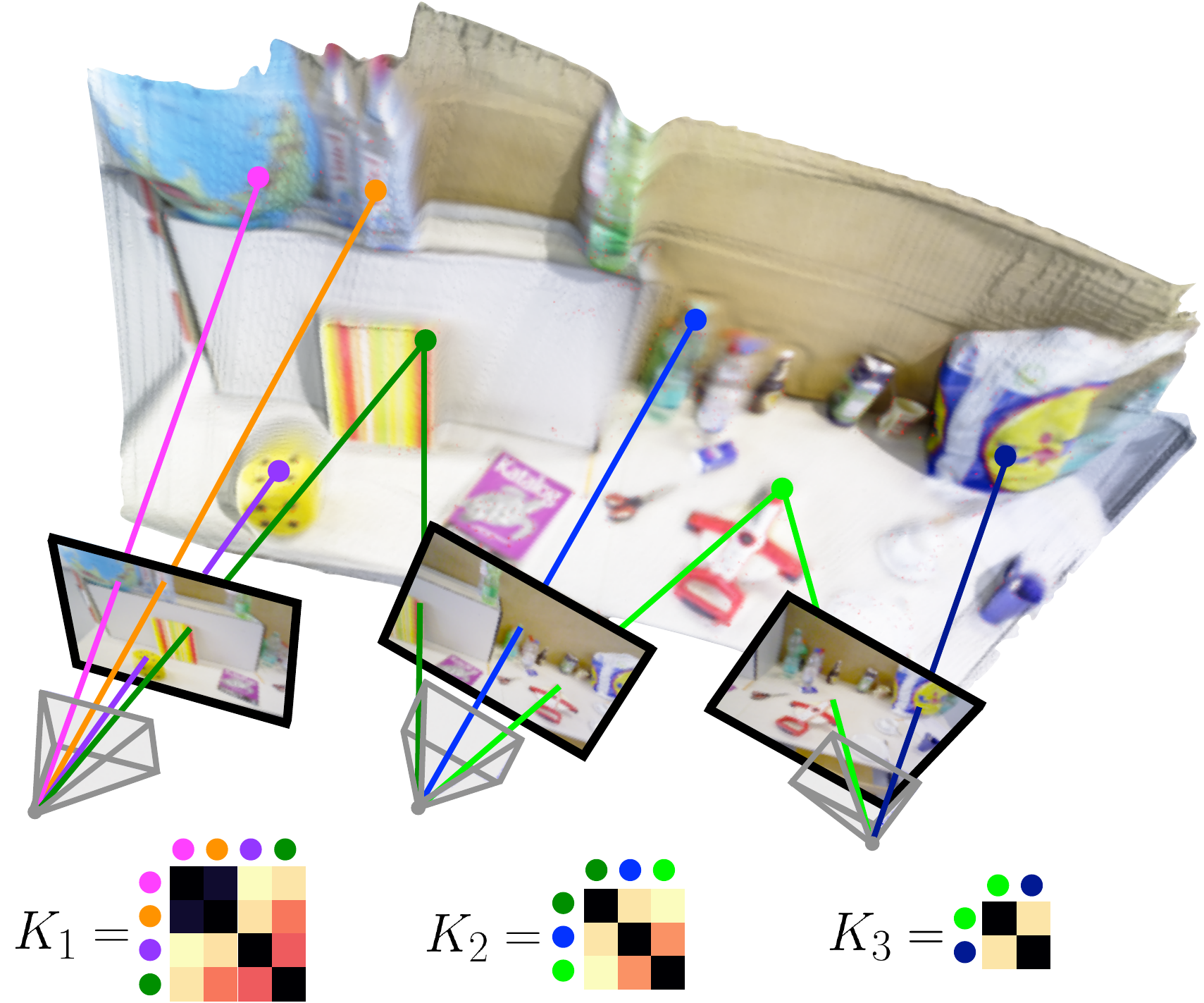}
	\caption{Example monocular reconstruction using the depth covariance for bundle adjustment and dense depth prediction from three seconds (100 frames) of TUM data.  Three representative images and the mesh from TSDF fusion of the depth predictions are shown. Each frame leverages the learned covariance function to model geometric correlation between pairs of scene points.}
	\label{fig:title_fig} 
\end{figure}

In this paper, rather than directly predicting geometry from images, we propose learning a depth covariance function. Given an RGB image, our method predicts how the depths of any two pixels relate. To achieve this, a neural network transforms color information to a feature space, and a Gaussian process (GP) models a prior distribution given these features and a base kernel function.  The distinction between image processing and the prior enables promoting locality and granting flexible capacity at test time.  Locality avoids over-correlating pixels on distinct structures, while adaptive capacity permits tuning the complexity of our subspace to the content of the viewed scene. 

Learning this flexible, high-level prior allows for balancing data-driven methods with test-time optimization that can be applied to a variety of geometric vision tasks. Furthermore, the covariance function is agnostic to the 3D representation as it does not directly learn a geometric output.  Depth maps may be requested by conditioning on observations, but the prior may also be leveraged for inferring the desired latent 3D representation. In Figure \ref{fig:title_fig}, we illustrate depth covariance along with an example of bundle adjustment, dense depth prediction, and multi-view fusion.

In summary, our key contributions are:
\begin{itemize}
    \item A framework for learning the covariance function by selecting a depth representation, a base kernel function, and a scalable optimization objective
    \item Application of the prior to depth completion, bundle adjustment, and monocular dense visual odometry
\end{itemize}

%% file: sections/related_work.tex
\section{Related Work}
\label{sec:related_work}

\noindent \textbf{View-based Priors for Geometric Inference}
Traditionally, sparse reconstruction methods have ignored per-image correlations \cite{schonberger_structure--motion_2016}, 
\cite{mur-artal_orb-slam2_2017}, while dense methods \cite{newcombe_dtam_2011}, \cite{engel_lsd-slam_2014} have used naive priors based on neighboring pixels.  With the availability of depth data and the rise of deep learning, a variety of learned depth priors have been proposed.  Although monocular depth estimation methods that directly predict geometry \cite{eigen_depth_2014}, \cite{godard_digging_2019}, \cite{ranftl_towards_2022} have demonstrated great progress, they may predict irreparable mistakes and do not allow for handling ambiguities.  Utilizing surface normals may refine these depth estimates \cite{bae_irondepth_2022}, but the improvement is still subject to errors from the initial depth prediction.  For this reason, we focus on approaches that balance learned priors and test-time optimization. Popular approaches include predicting low-dimensional latent codes \cite{bloesch_codeslam_2018}, \cite{czarnowski_deepfactors_2020} and generating a basis of depth maps \cite{tang_ba-net_2019}, \cite{graham_ridgesfm_2020}.  While these methods utilize single image information to reduce the dimensionality of depth map inference, they fix capacity during training, produce overly smooth depth maps, and create over-correlated global changes of the depth map.  Predicting mesh vertices in the image plane \cite{bloesch_learning_2019} permits distribution of capacity and optimizing depth, but the number of latent variables is still fixed and an explicit mesh representation cannot easily model complex scenes.  

Instead of predicting geometry directly, we focus on predicting geometric correlation.  By decoupling the pipeline into an image processing network and a GP, the network is not responsible for learning geometry directly and we do not need to fix the geometric capacity.  The dimensionality of the subspace may be adapted for representing low-rank scenes more compactly and complex geometry with high-fidelity.  Lastly, we use a covariance function with locality to bias learning towards local appearance information.

\noindent \textbf{Learning Covariance Functions} 
Choosing covariance functions and performing model selection for GPs is a well-studied topic \cite{rasmussen_gaussian_2005}.  While stationary kernels have been explored for merging LiDAR observations with monocular depth estimates \cite{yoon_balanced_2020}, more expressive kernels are required for sparse observations. Nonstationary covariance functions using local Gaussian parameterizations \cite{paciorek_nonstationary_2003} have shown potential in robotic terrain mapping \cite{lang_adaptive_2007}.  However, since GP model selection occurs per data example, optimizing hyperparameters is challenging, and novel nonstationary kernels have been proposed to limit flexibility \cite{chen_ak_2022}.

Rather than performing model selection per image, we are inspired by deep kernel learning (DKL) \cite{wilson_deep_2016}, which leverages training data to predict kernel hyperparameters.  Alternatively, this may be viewed as meta-learning, where each task is an RGB-depth example \cite{patacchiola_bayesian_2020}.  To mitigate over-correlation when using stationary kernels in feature space as proposed in DKL \cite{ober_promises_2021}, we utilize local information in pixel space \cite{paciorek_nonstationary_2003} to model surface geometry.  While deep nonstationary kernel regression has been explored for depth completion \cite{liu_learning_2021}, GPs balance data fit and model complexity during training, and the uncertainty estimates are conducive to decision-making and inference in optimization frameworks.

\noindent \textbf{Residual Covariances in Computer Vision}
Uncertainties in machine learning are often divided between two types: model uncertainty and data uncertainty \cite{kendall_what_2017}.  For vision problems, model uncertainty is often ignored due to the availability of data and for tractable optimization.  Residual uncertainty is usually predicted from a network as a per-pixel variance due to high-dimensionality, but the assumption of independence ignores the correlation present in images.  For example, variational autoencoders (VAEs) \cite{kingma_autoencoding_2014} with a diagonal likelihood output overly smooth mean predictions and unnatural samples with salt-and-pepper noise.  To introduce correlation in the likelihood, structured uncertainty prediction networks (SUPNs) predict a full information matrix by defining a graph topology consisting of specific neighbors \cite{dorta_structured_2018}.  Since this matrix is sparse, it can be efficiently inverted to obtain the covariance matrix.  This has also been leveraged to distill monocular depth prediction ensembles into a single SUPN \cite{simpson_learning_2022}.  

Similar to SUPNs, we move beyond diagonal covariance approximations that lack correlation.  However, we learn a covariance function which does not require a predefined graph topology and allows for long-range correlations beyond neighboring pixels.  Furthermore, the marginal distribution of any set of variables may be examined without constructing the full joint distribution, which is of great interest for compact inference in geometric vision.

%% file: sections/learning_depth_cov.tex
\section{Learning a Depth Covariance Function}
\label{sec:learning_depth_cov}

Given an image, we model a distribution over possible depth functions via a Gaussian process (GP) \cite{rasmussen_gaussian_2005}.  With the input domain being normalized image coordinates $\mathbf{x}_i = (u_i,v_i)$ where $u_i,v_i \in [-1, 1]$, the outputs for any finite set of inputs are jointly Gaussian. The GP is then defined by a mean function $m(\mathbf{x})$ and a covariance function $k(\mathbf{x}, \mathbf{{x}'})$:
\begin{align}
    f(\mathbf{x}) \sim \mathcal{GP}\left(m(\mathbf{x}),\, k(\mathbf{x}, \mathbf{{x}'}) \right) .
\end{align}
For learning the parameters of this GP, we need to select the representation of depth, the mean and covariance functions, and the optimization objective.  We will outline our decisions in the following sections, but there exist many ways to define this prior.  

\subsection{Depth Representation}
\label{subsec:depth_rep}

When predicting depth, we often want to attenuate errors further from the camera view, so that nearby structures are prioritized.  Examples of such representations include inverse depth \cite{civera_inverse_2008}, disparity \cite{ranftl_towards_2022}, and log-depth \cite{eigen_depth_2014}. We select log-depth for two reasons.  First, the log-normal distribution is more suitable for skewed distributions \cite{rendu_normal_1979}, as the range of the GP is unbounded.  Representing depth or inverse depth with a normal distribution would require \textit{ad hoc} truncation, as depth functions could move behind the camera.   Second, we represent depth with relative scale, as the absolute scale ambiguity in monocular images accounts for much of the error in learned depth prediction \cite{eigen_depth_2014}.  We can then focus on learning the relationship between predictions.  In the log-depth formulation, a constant mean $m(\mathbf{x}_i) = m$ directly corresponds to a scale variable.  Given a log-depth observation $\mathbf{y}_i = \log{\mathbf{d}_i}$, we can adjust its scale via this mean:
\begin{align}
    e^{\mathbf{y}_i - m} = e^{\mathbf{y}_i} e^{-m} = C e^{\mathbf{y}_i} = C\mathbf{d}_i.
\end{align}
In other words, the median of the depth distribution is controlled by the mean of the log-normal distribution.  During training, similar to \cite{eigen_depth_2014}, we may find the optimal scale $m$ that minimizes our data loss. In addition, the log-depth representation allows covariance to be in relative scale.  At test time, the scale may be fixed if known, or jointly optimized with other variables.  Learning a more expressive mean function per image would be useful in the absence of any depth observations, but introduces many degrees of freedom.  For GPs, generalization is largely dependent on the covariance function, so we leave the mean to be a constant per image.

\begin{figure}[t]
    \includegraphics[width=\columnwidth]{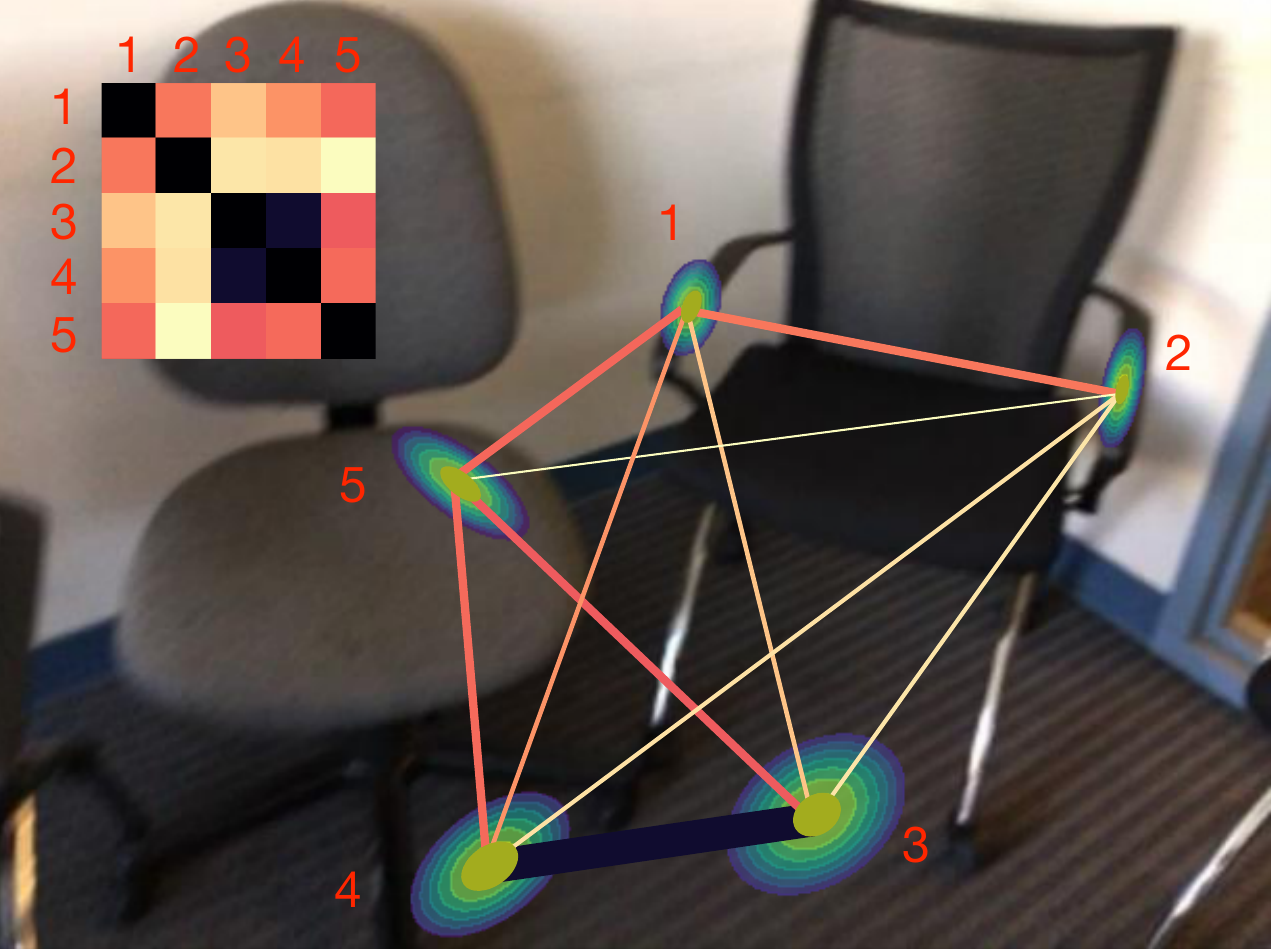}
    \caption{Visualizing our depth covariance function: for every pixel of an input image, the trained network predicts a 2D kernel matrix. Here we show the covariance function between pairs of pixels in both matrix form and as edges in a graph, with the line thickness representing the magnitude of covariance.}
    \label{fig:method_fig} 
\end{figure}

\subsection{Covariance Function}
\label{subsec:cov_func}

We would like to learn the parameters of a depth covariance function using pairs of RGB and depth images.  By definition, the covariance function must be positive semidefinite (PSD).  In our framework, an RGB image is fed into a convolutional neural network (CNN), which outputs features for a base covariance function.  We model the depth observations as being jointly Gaussian so that the CNN and base kernel hyperparameters may be jointly learned.

For the base covariance function, we would like to avoid over-correlation of independent structures.  To achieve local influence, we use the family of nonstationary kernels described in \cite{paciorek_nonstationary_2003}.  Each point $\mathbf{x}_i$ in an image can be viewed as a Gaussian distribution with a 2D kernel matrix $\Sigma_i$.  Then, the covariance between two points is the convolution of these two densities over the input domain with a normalizing constant.  Alternatively, this can be viewed as the similarity between distributions via the Bhattacharyya kernel \cite{jebara_probability_2004}.  Given an isotropic covariance function $R^S(\sqrt{q})$, the closed-form expression is:
\begin{align}
    k(\mathbf{x}_i, \mathbf{x}_j) &= \sigma_f^2 \frac{|\Sigma_i|^{1/4} |\Sigma_j|^{1/4}}{|\Sigma_i + \Sigma_j|^{1/2}} R^S(\sqrt{q}), \\
    q &= (\mathbf{x}_i-\mathbf{x}_j)^T (\Sigma_i + \Sigma_j)^{-1} (\mathbf{x}_i-\mathbf{x}_j),
\end{align}
where $\sigma_f^2$ is a learnable signal variance.  To better handle discontinuities, we select the Mat\'{e}rn function as our isotropic covariance function.  With the base covariance function requiring a 2D PSD matrix for each pixel, the CNN outputs three channels $(c_1, c_2, c_3)$, which are parameterized given the positive diagonal and determinant constraints:
\begin{align}
    \Sigma_i &= \begin{bmatrix} e^{c_1} & \tanh(c_3) \sqrt{e^{c_1}*e^{c_2}} \\ 
                    \tanh(c_3) \sqrt{e^{c_1}*e^{c_2}} & e^{c_2}  \end{bmatrix}.
\end{align}
An example of five kernel matrices and the marginal covariance in matrix and graph form is shown in Figure \ref{fig:method_fig}. We use a UNet architecture \cite{ronneberger_u-net_2015} and output features at different levels for a multi-scale loss.  Each scale also has its own GP hyperparameters, signal variance $\sigma_f^2$ and noise variance $\sigma_n^2$.  Depth maps are coarsened for lower-levels, so finer levels are weighted higher during the total loss calculation.

While a more global covariance function, such as a squared-exponential over output features, would provide additional flexibility, we avoid this for two reasons.  First, changes in depths on one part of the scene may have significant influence on a completely unrelated part, as seen in \cite{bloesch_codeslam_2018}.  Second, deep kernel learning using this setup is unstable and biased towards over-correlating the input domain \cite{ober_promises_2021}.   By directly using pixel coordinates in our covariance function, we restrict changes in geometry to be local and bias the network to learn local appearance information.  

\subsection{Optimization Objective}
\label{subsec:learning}

In GP literature, model selection is performed by minimizing the negative log marginal likelihood (NLML)
\begin{align}
    -\log p(\mathbf{y} | X) &= \frac{1}{2} (\mathbf{y} - m)^T (K_{\text{ff}} + \sigma_n^2 I)^{-1} (\mathbf{y} - m) \notag\\
    &+ \frac{n}{2} \log 2\pi + \frac{1}{2} \log |K_{\text{ff}} + \sigma_n^2 I|,
\end{align}
where the matrix $K_{\text{ff}}$ is a PSD matrix with entries defined by the covariance function.  The first term is often referred to as the ``data fit'', while the third term is the ``complexity penalty''.  In our scenario, the complexity term is minimized by correlating points, while the data term ensures the ground-truth depth map is plausible given the mean and predicted covariance.

However, the marginal likelihood for a GP requires a $O(N^3)$ matrix inversion.  For pixels in an image, this becomes intractable.  Since depth images are relatively low-rank, we use a sparse GP approximation \cite{bauer_understanding_2016} during training by randomly sampling inducing points for a Nystr{\"o}m approximation to the full covariance matrix.  Given the covariance matrix for the inducing points $K_{\text{uu}}$ and the cross-covariance between the inducing points and the entire domain $K_{\text{uf}}$, we have:
\begin{align}
    K_{\text{ff}} \approx \tilde{K}_{\text{ff}} \triangleq K_{\text{fu}} K_{\text{uu}}^{-1} K_{\text{uf}} .
\end{align}
Specifically, we use the variational free energy (VFE) framework \cite{titsias_variational_2009}, which defines our training loss as
\begin{align} \label{eq:vfe_loss}
    \mathcal{F} &= \frac{1}{2} (\mathbf{y} - m)^T (\tilde{K}_{\text{ff}} + \sigma_n^2 I)^{-1} (\mathbf{y} - m) + \frac{n}{2} \log 2\pi \notag\\
    &+ \frac{1}{2} \log |\tilde{K}_{\text{ff}} + \sigma_n^2 I| + \frac{1}{2\sigma_n^2} \text{tr}{(K_{\text{ff}}-\tilde{K}_{\text{ff}})}.
\end{align}
The first three terms are the same as the original NLML, but using the approximate covariance, while the last term penalizes the conditional variances at all inputs given the inducing points, which only requires the diagonal of the full covariance matrix.  Note that we also assume homoscedastic observation noise variance $\sigma_n^2$ across the entire dataset.

%% file: sections/leveraging_depth_cov.tex
\section{Leveraging a Depth Covariance Function}
\label{sec:leveraging_depth_cov}

Given an image and its corresponding CNN output maps, we may leverage the covariance function by defining a prior over depth functions, conditioning on depth observations to yield a predictive distribution, and actively sampling pixel locations that minimize the variance of depth predictions.

\subsection{Gaussian Process Prior}
\label{subsec:gp_prior}

Given a finite set of input points $X_*$, we may define a Gaussian prior over their log-depths:
\begin{align}
    \mathbf{f}_* \sim \mathcal{N}\left(m,\, K(X_*, X_*) \right) .
\end{align}
The GP defines a prior over functions, and for any number of input points, we obtain a Gaussian prior.  For geometric vision tasks, this prior can be leveraged for anything from sparse to dense methods.  Due to the marginalization property of GPs, adding points to the indexing set will not change the marginal distribution of the existing set.  Alternatively, the prior may be viewed as an image-conditioned regularizer of any desired latent geometry parameterization.

\subsection{Predictive Distribution}
\label{subsec:conditional}

\begin{figure}[t]
	\centering
	\includegraphics[width=\columnwidth]{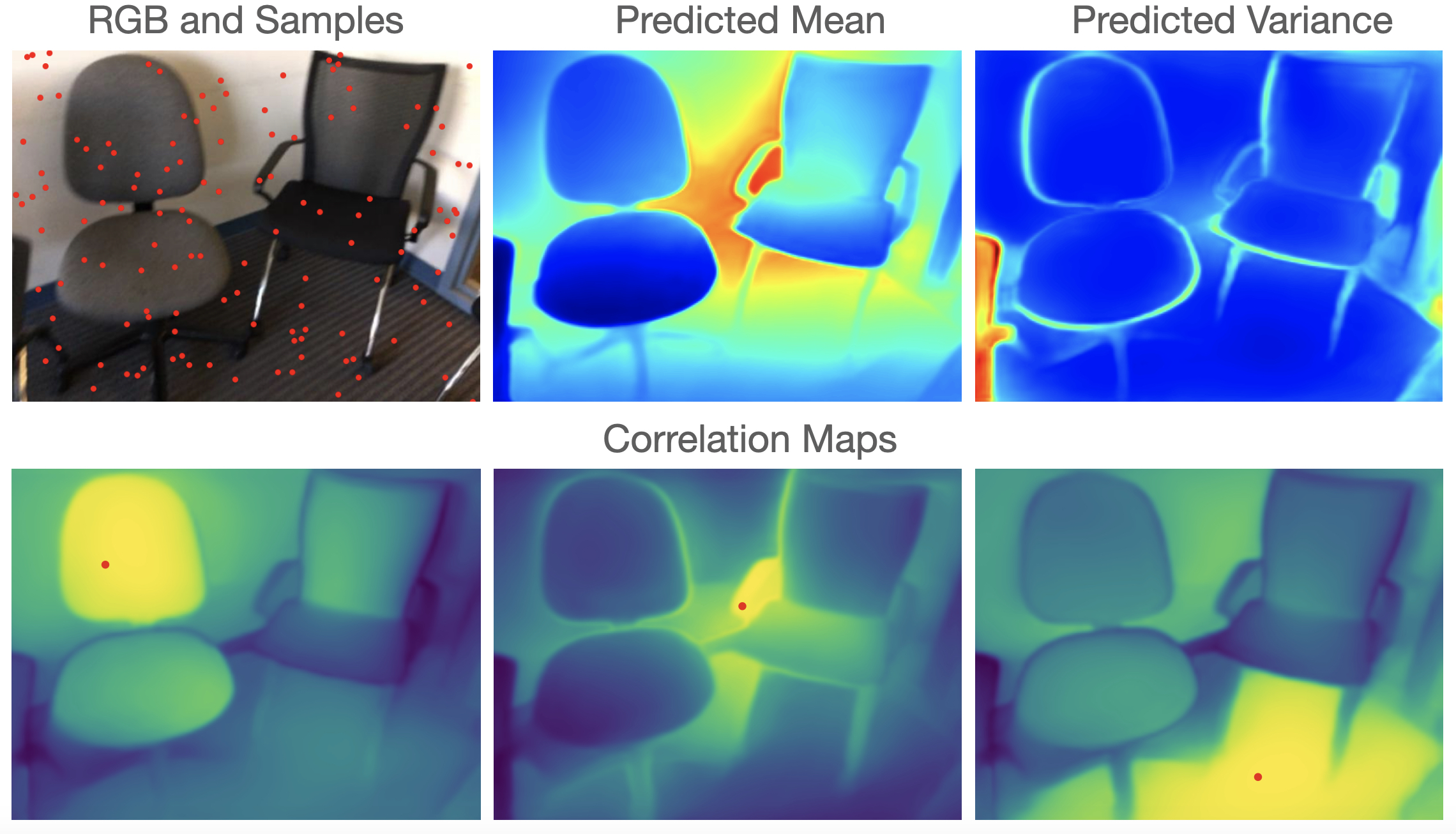}
	\caption{Conditioning example for 128 samples.  The posterior variance is high around edges and in areas lacking samples.  The columns of $K_{\text{fn}}$, or correlation maps, are shown for select points.}
	\label{fig:conditioning} 
\end{figure}

In many cases, log-depth observations $\mathbf{y}$, such as from RGB-D sensors or LiDAR, may be available.  We can explicitly condition the prior on these observations to obtain a posterior distribution.  The predictive mean $\mathbf{f}_*$ and covariance $\Sigma_*$ given $N$ samples is:
\begin{align}
    \mathbf{f}_* &= m + K_{\text{fn}} (K_{\text{nn}} + \sigma_n^2 I)^{-1} (\mathbf{y} - m), \\
    \Sigma_* &= K_{\text{ff}} - K_{\text{fn}} (K_{\text{nn}} + \sigma_n^2 I)^{-1} K_{\text{nf}}.
\end{align}
The predictive mean is a linear function in terms of the observations $\mathbf{y}$, so test-time inference of latent depths is efficient.  If a full covariance is not required, only certain blocks or per-pixel variances need to be calculated. Note that the covariance depends only on the observation coordinates and not the observed values.  We visualize the predictive mean, variance, and correlation maps for an example in Figure \ref{fig:conditioning}.

\subsection{Active Point Selection}
\label{subsec:active_selection}

For dense reconstruction, it is beneficial to construct a compact representation that can achieve high-fidelity results.  As mentioned previously, the predictive covariance depends only on the RGB image and locations of depth observations, but not on the observations themselves.  By viewing the CNN as a meta-learned initialization of the nonstationary kernel parameters, we may use 2D observations as a proxy for the complexity of 3D geometry.  Active selection of informative pixels up to a desired variance permits representing less complex scenes with fewer samples and allocating capacity towards high-frequency geometry. 

Inspired by sensor placement literature \cite{guestrin_near-optimal_2005}, an entropy-based criterion is used to select informative pixels.  In the greedy-case, this simplifies to selecting the input point with the highest conditional variance at each step.  Since this requires computing the conditional variance for each newly added point, we leverage incremental updates to the variance and Cholesky factorization of the training covariance matrix \cite{ranganathan_online_2011}.  By decoupling the neural network and GP, we do not require any additional network passes.

\begin{figure}[t]
	\centering
	\includegraphics[width=0.99\columnwidth]{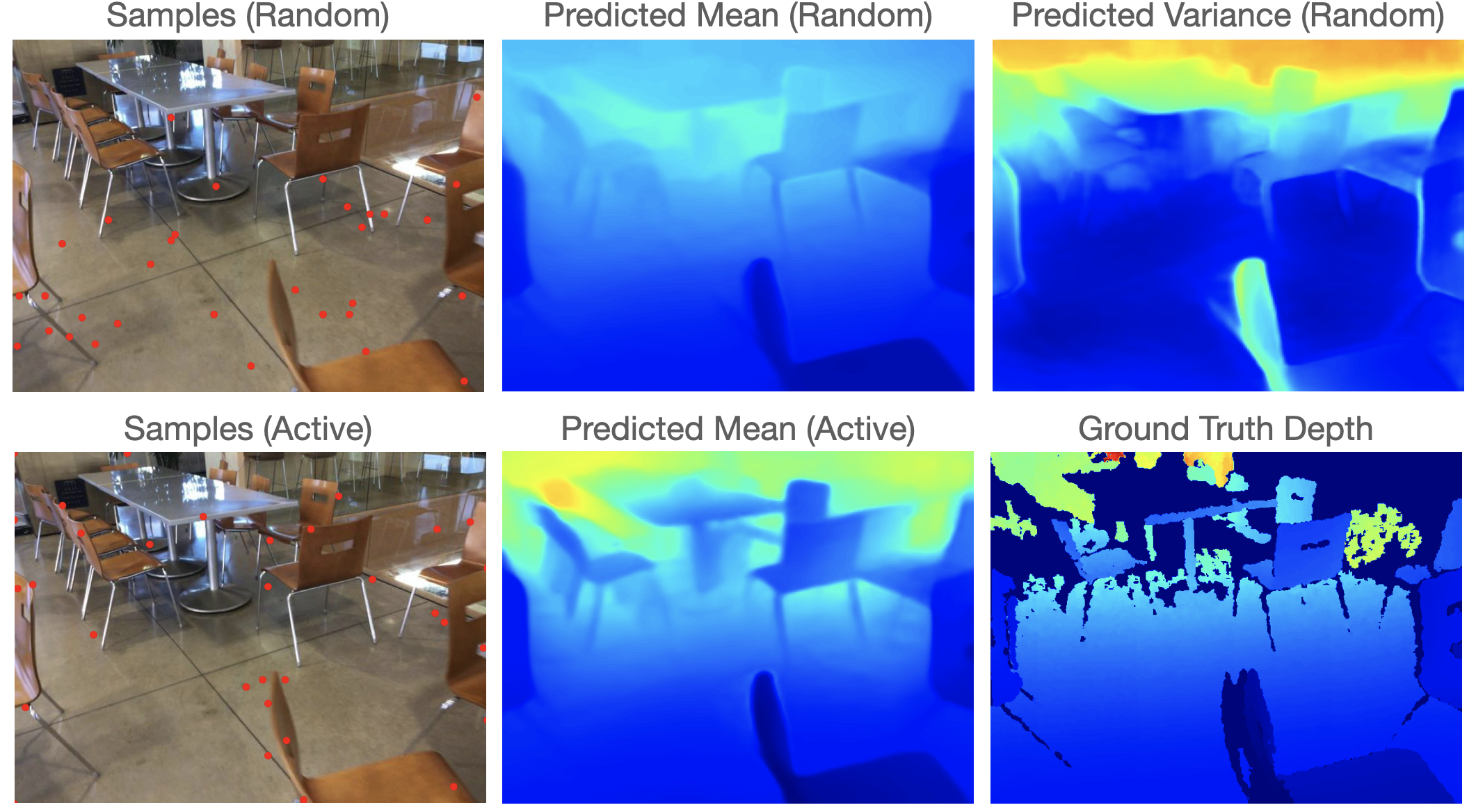}
	\caption{Qualitative comparison of random and active sampling given 32 sample selections.  Random sampling misrepresents depth at the top of the image, while active sampling focuses on the chair geometry and avoids redundant samples on the floor.}
	\label{fig:qual_sampling} 
\end{figure}

A qualitative example of active sampling is shown in Figure \ref{fig:qual_sampling}.  Random sampling severely misrepresents depth of table and chairs, which all appear at similar depths.  Active sampling focus samples around the thin chair edge near the image, while also avoiding oversampling on the floor.  Furthermore, active sampling avoids missing the top section of the image, so that the table and chairs are well-represented.

%% file: sections/applications.tex
\section{Applications}
\label{sec:results}

We apply depth covariance to three fundamental geometric vision tasks: depth completion, bundle adjustment, and monocular dense visual odometry (DVO).  For depth completion, we use the NYUv2 benchmark \cite{silberman_indoor_2012} and the train-test splits from \cite{ma_sparse_2018}.  For bundle adjustment and DVO, we train the covariance function on ScanNet \cite{dai_scannet_2017}, and evaluate on the TUM RGB-D dataset \cite{sturm_benchmark_2012}.  For the UNet, we use an input size of $256 \times 192$, 16 channels after the first layer, 5 downsampling steps, and 4 output levels.  This results in roughly 9 million parameters. 

\subsection{Depth Completion}
\label{subsec:depth_completion}

Depth completion is a fundamental task that will be leveraged for additional applications. We may directly condition on sparse observations as described in Section \ref{subsec:conditional} to obtain a dense depth map and uncertainties.

\subsubsection{Completion Accuracy and Error}

Since our method does not predict a specific instance of geometry, we report errors with respect to the GP posterior mean.  We compare to the foundational Sparse-to-Dense (S2D) \cite{ma_sparse_2018}, deep kernel regression with (KernelNet+R) and without refinement (KernelNet) \cite{liu_learning_2021}, and a recent state-of-the-art network RigNet \cite{yan_rignet_2022}. Results on the validation set with 500 random samples are shown in Table \ref{tab:depth_comp_nyuv2}.

\setlength{\tabcolsep}{2pt}
\begin{table}[h]
    \centering
    \begin{tabular}{|l|c|c c c c c|}
        \hline
        \multirow{2}{*}{Method} &
        \multicolumn{1}{c}{Error ($\downarrow$)} &
        \multicolumn{5}{|c|}{Accuracy ($\uparrow$)}
        \\
        & RMSE & $\delta_{1.02}$ & $\delta_{1.05}$ & $\delta_{1.10}$ & $\delta_{1.25}$ & $\delta_{1.25^2}$ \\
        \hline 
        S2D \cite{ma_sparse_2018} & 0.204 & - & - & - & 97.8 & 99.6 \\
        KernelNet \cite{liu_learning_2021} & 0.198 & 65.5 & 82.9 & 91.6 & 97.7 & 99.8 \\
        KernelNet+R \cite{liu_learning_2021} & 0.111 & 84.8 & 94.1 & 97.4 & 99.3 & 99.9 \\
        RigNet \cite{yan_rignet_2022} & 0.090 & - & - & - & 99.7 & 99.9 \\
        Ours & 0.157 & 81.2 & 92.6 & 97.1 & 99.4 & 99.9\\
        \hline
    \end{tabular}
    \caption{Depth completion on NYUv2 with 500 sampled points.}
    \label{tab:depth_comp_nyuv2}
\end{table}

KernelNet also predicts three feature maps for 2D covariance parameters, but we achieve  better performance under the GP framework.  We achieve comparable \textit{accuracy} to state-of-the-art methods KernelNet+R and RigNet while using fewer parameters. We also do not convert the problem into classification \cite{liu_learning_2021} or have complex forward passes with iterative layers \cite{yan_rignet_2022}.  The depth covariance outperforms methods with similar UNet architectures in RMSE.  While the \textit{error} is not state-of-the-art, we use a lightweight network and do not train specifically for the single task of depth completion with the number of samples known \textit{a priori}.

We also explore varying sparsity, as depth completion networks are trained for a specific number of samples.  A comparison of RMSE for a varying number of test samples is shown in Table \ref{tab:depth_comp_nyuv2_sparsity}.  Depth covariance outperforms traditional depth completion methods for sparse inputs, and is competitive with SpAgNet \cite{conti_sparsity_2023} which is designed specifically for these cases and contains an additional non-local spatial propagation layer.  We note that for \textit{error} metrics, we simply use the posterior mean for comparison.  However, the GP provides a distribution over depths, not just a single instance, and we will demonstrate additional capabilities such as active decision-making and inferring latent geometry when direct observations are not present.

\setlength{\tabcolsep}{5pt}
\begin{table}[h]
    \centering
    \begin{tabular}{|l|r|r|r|r|}
        \hline
        \# Samples & 5 & 50 & 100 & 200 \\
        \hline
        CSPN \cite{cheng_depth_2018} & 2.063 & 0.884 & 0.388 & 0.177 \\
        NLSPN \cite{park_nonlocal_2020} & 1.033 & 0.423 & 0.246 & 0.\textbf{142} \\
        SpAgNet \cite{conti_sparsity_2023} & \textbf{0.467} & \textbf{0.272} & \textbf{0.209} & \underline{0.155} \\ 
        Ours & \underline{0.717} & \underline{0.298} & \underline{0.236} & 0.193 \\
        \hline
    \end{tabular}
    \caption{Depth completion RMSE (m) on NYUv2 with a varying number of input samples. The best result is in bold, while the second best is underlined.}
    \label{tab:depth_comp_nyuv2_sparsity}
\end{table}

\subsubsection{Posterior Uncertainty}

Calibrated uncertainties are beneficial for balancing multiple views and sensors in optimization.  Under-confidence does not account for the full information given by the constraints, while over-confidence may bias the solution.  Given that the GP provides uncertainties in addition to the mean, we evaluate the calibration properties for the depth completion setup from Section \ref{subsec:depth_completion}.

\begin{figure}[t]
	\centering
	\includegraphics[width=0.87\columnwidth]{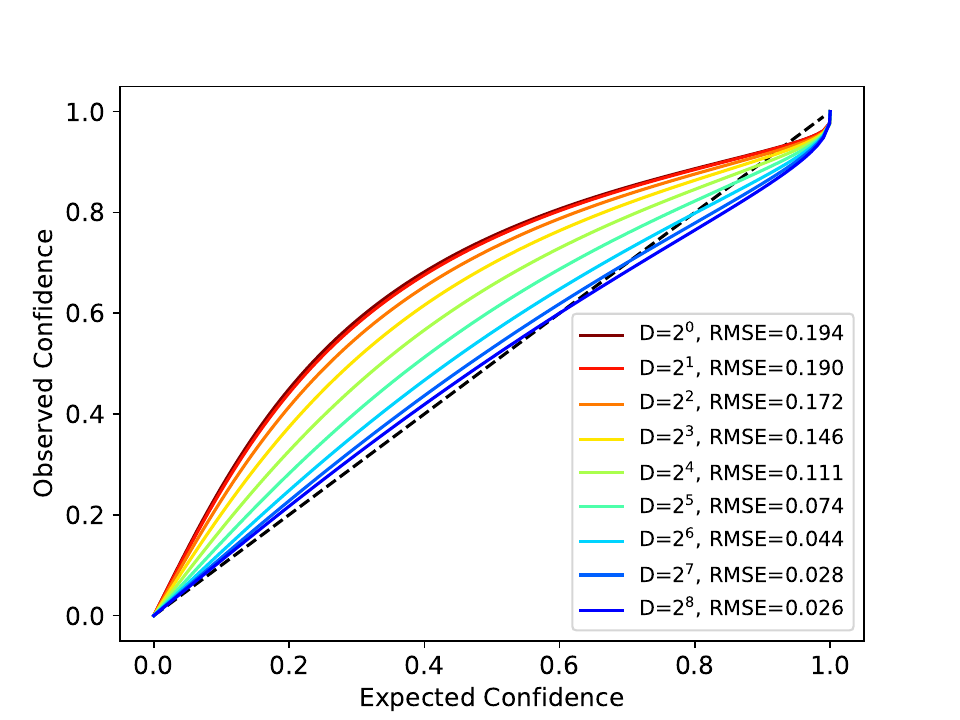}
	\caption{Calibration plots of varying posterior marginal covariance block dimensions $D$ on NYUv2 depth completion.  The ideal calibration is $y=x$, where the observed confidence matches expected confidence.  The region above the line indicates model under-confidence, while the area below signals over-confidence.}
	\label{fig:calib} 
\end{figure}

Since most methods typically estimate per-pixel observation noise, we assess the effect of including off-diagonal terms.  Calibration plots measure how well the model's predicted confidence matches the expected confidence.  To extend calibration plots for regression beyond variance \cite{kuleshov_accurate_2018}, we extract marginal covariances of varying dimension $D$ and calculate Mahalanobis distances using this block-diagonal approximation to the full covariance matrix. We then count the frequency of distances below varying chi-square thresholds that define the expected confidence.  In the variance case with $D=1$, residuals do not affect each other.  We plot the results in Figure \ref{fig:calib} along with the RMSE against the ideal calibration, where the empirical confidence matches the observed confidence.  Including more off-diagonal terms improves the model calibration, and significantly reduces the under-confidence of the model.

\subsubsection{Active Sampling Evaluation}
\label{subsec:active_eval}

\begin{figure}[t]
	\centering
	\includegraphics[width=\columnwidth]{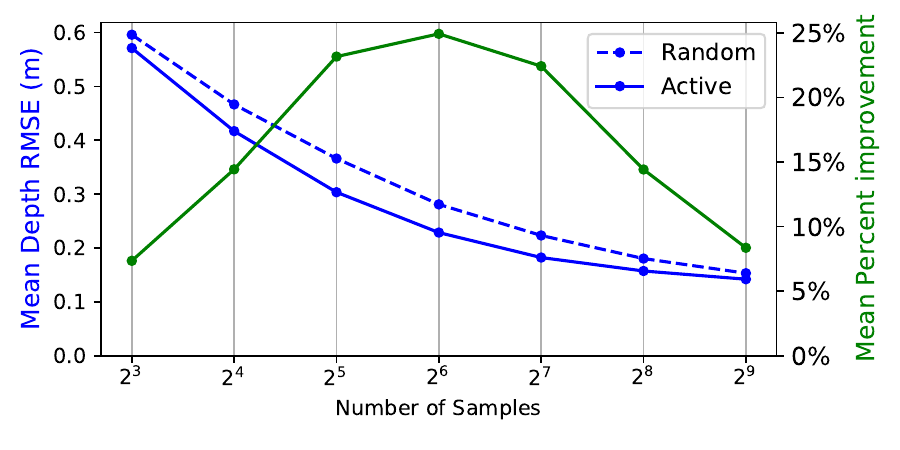}
	\caption{Mean depth completion RMSE for random and active sampling with a varying number of samples.  Green line shows mean percent improvement of active over random sampling.}
	\label{fig:sampling} 
\end{figure}

While the location of depth samples is often not a free variable, such as from LiDAR or depth sensors, in monocular vision, we may wish to actively estimate the depths of certain pixels to improve downstream tasks such as dense reconstruction.  We investigate whether the meta-learned covariance parameters may be leveraged for selecting more informative pixels.  We compare randomly sampling pixels uniformly against the greedy conditional variance as described in Section \ref{subsec:active_selection}.  The effect on depth error and the mean percentage improvement for a varying number of samples is shown in Figure \ref{fig:sampling}.  Note that the greedy active sampling consistently outperforms random sampling.  For a large number of samples, the relative improvement of active sampling decreases as there is a sufficient number of observations.  For a few samples, only coarse depth structure is retained, and the greedy selection occasionally encourages samples near the image boundaries.  Beyond greedy selection, other sampling methods that explicitly minimize uncertainty over the whole domain may demonstrate improved performance at the expense of computational cost.

\subsection{Bundle Adjustment}
\label{subsec:bundle_adjustment}

Bundle adjustment is the foundation for many vision pipelines, and when successful, it produces very accurate camera poses.   However, the standard formulation of assuming independence between observations proves challenging for monocular SLAM systems, which may fail during initialization or when translation is negligible compared to rotation.  Many realistic scenarios require robust and fast initialization with little motion.  Therefore, we evaluate the use of the depth covariance in small baseline scenarios.  We also show that the depth covariance is not limited to 2D depth map inference, but can also be used to infer 3D structure.

\begin{figure}[t]
	\centering
	\includegraphics[width=0.95\columnwidth]{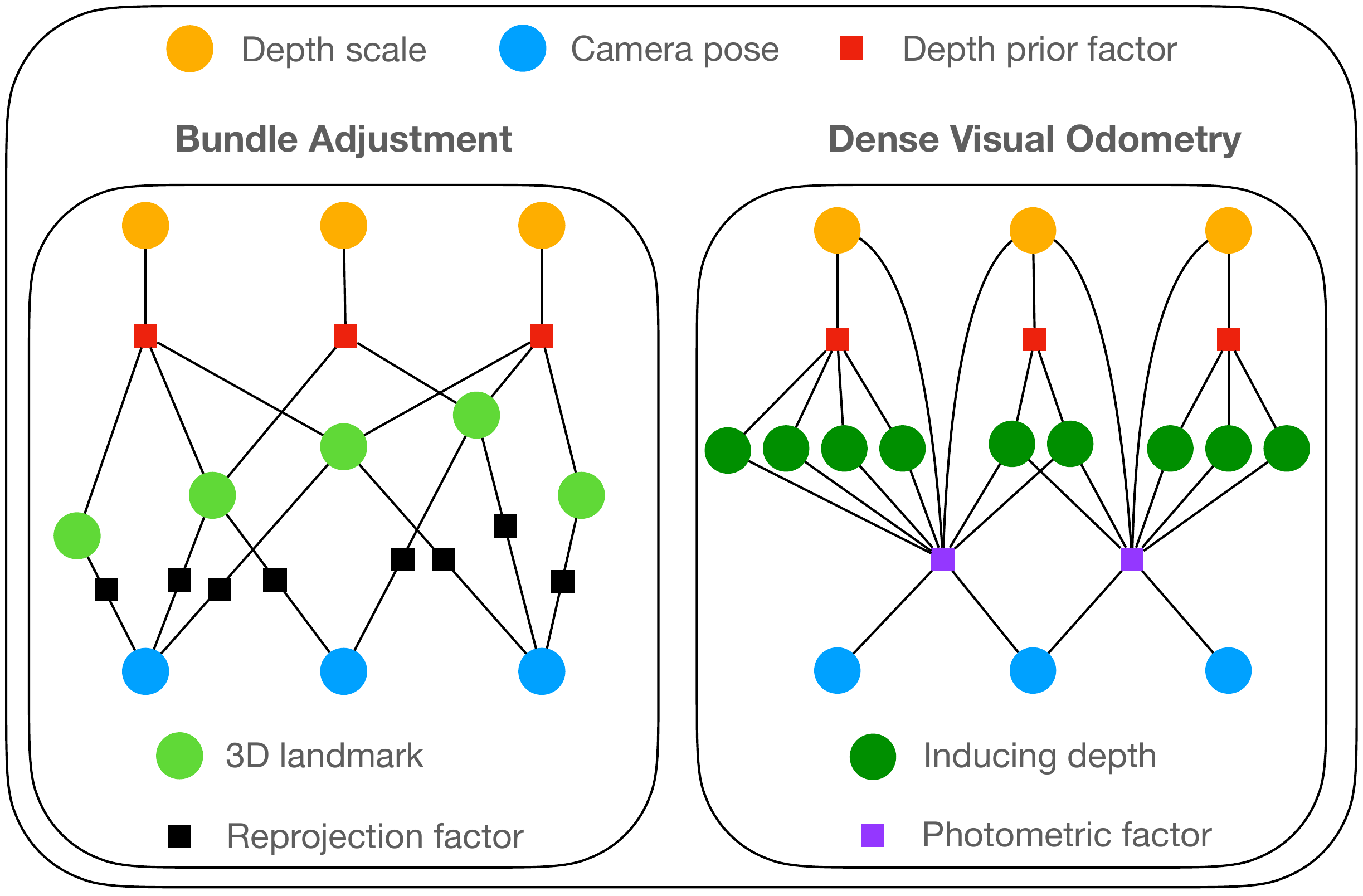}
	\caption{Factor graphs for bundle adjustment and monocular dense visual odometry.  Pose and scale priors to constrain gauge freedom are omitted for simplicity.}
	\label{fig:ba_dvo_fg} 
\end{figure}

Bundle adjustment jointly optimizes camera poses and point landmarks given pixel correspondences.  Traditionally, the cost involves the sum of reprojection errors, as well as pose and scale priors on the first pose to fix gauge freedom which we omit for brevity.  We also add our depth prior factor per camera, so for landmarks $\mathbf{P}$ in the world frame, and poses $T_{cW}$ for $c = 1,...,C$, we have:
\begin{align}
    E &= \sum_c \sum_i || \pi(\mathbf{T}_{cW}, \mathbf{P}_i) - \mathbf{x}_{c,i} ||^2_{\sigma_r^2 I} \notag\\
    &+ \sum_c || \log \left( [T_{cW} \mathbf{P}_c]_z \right) - m_c ||_{K_{c}} ^2,
\end{align}
where each frame also has a scale variable $m_c$ and depth covariance $K_c$ for vectorized observed landmarks $\mathbf{P}_c$ in that frame.  The projection function $\pi$ projects 3D landmarks into the image plane.  To reduce the influence of outliers, we use the Huber cost function for reprojection errors.  The least-squares cost is optimized using Levenberg-Marquardt via GTSAM \cite{dellaert_gtsam_2022}.  The corresponding factor graph is shown on the left of Figure \ref{fig:ba_dvo_fg}.

\begin{figure}[t]
	\centering
	\includegraphics[width=0.95\columnwidth]{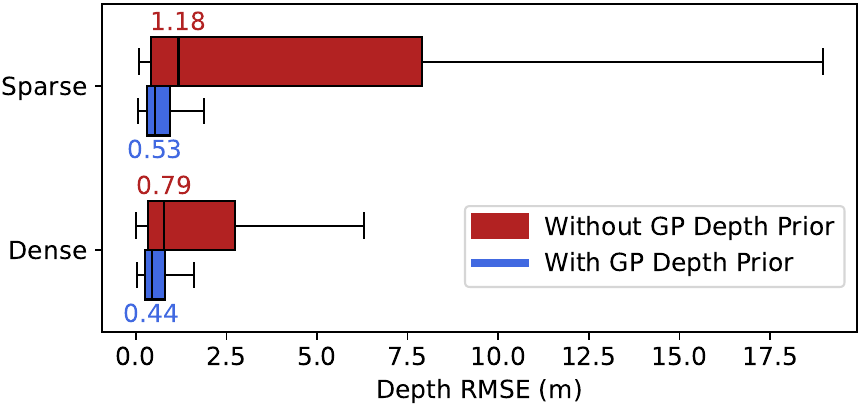}
	\caption{Boxplot of sparse and dense depth errors across all 5-frame windows, with and without the GP depth prior included in the optimization.  The median RMSE is written next to the plots.}
	\label{fig:bundle_adj_results} 
\end{figure}

\begin{figure}[t]
	\centering
	\includegraphics[width=0.95\columnwidth]{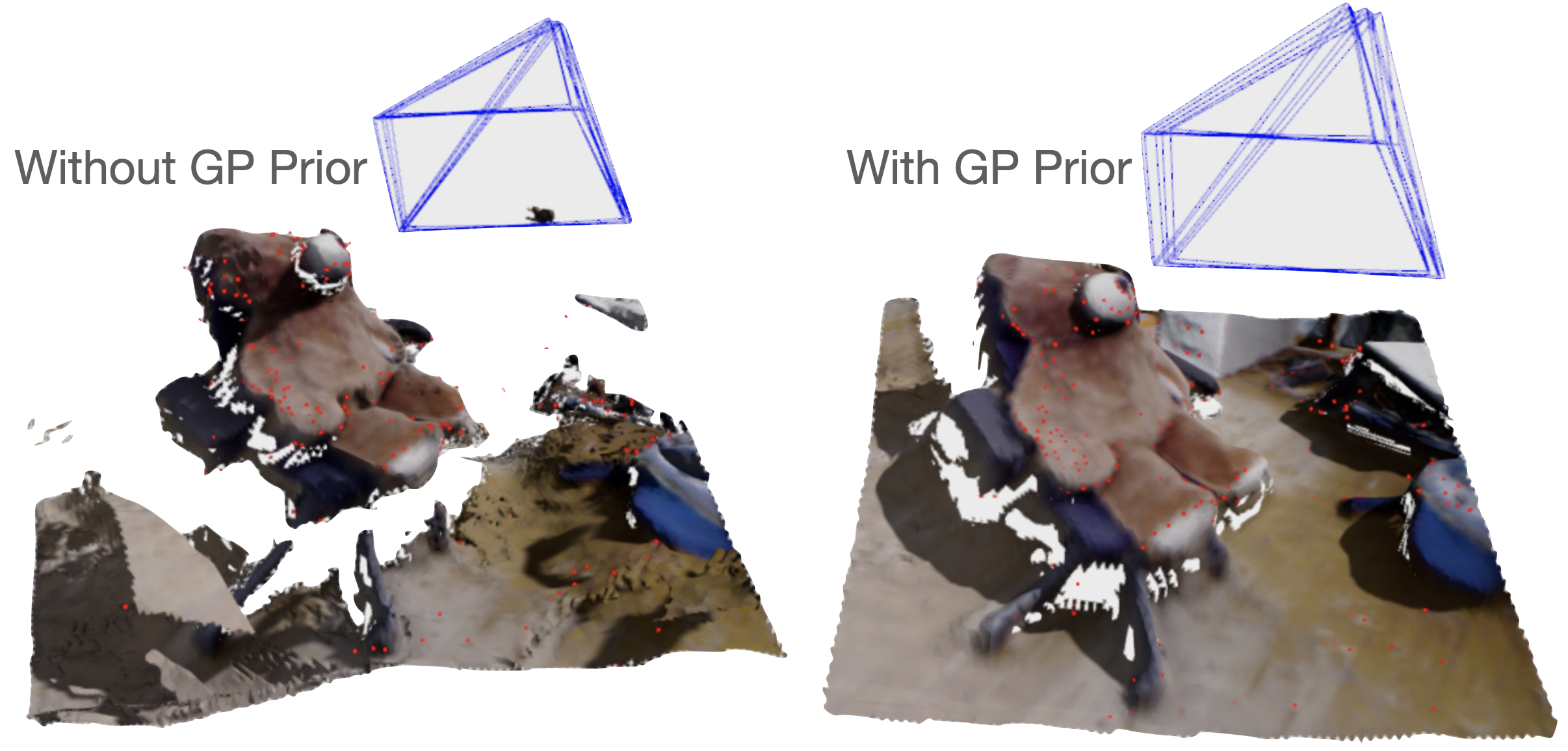}
	\caption{Qualitative example of small baseline bundle adjustment with and without the GP depth factor.  Both examples fuse 5 predicted depth maps densified from the optimized landmarks into a TSDF, and are visualized with identical lighting.}
	\label{fig:qual_ba_fig} 
\end{figure}

We divide all of the Freiburg 1 sequences from the TUM RGB-D dataset into small baseline 5-frame windows.  This yields 1544 sequences, emulating initializations to visual odometry systems.  For the frontend, Shi-Tomasi corners \cite{shi_good_1994} are detected and tracked using Lucas-Kanade tracking \cite{lucas_iterative_1981}, and outliers are filtered out via essential matrix RANSAC.  Poses are initialized using the motion capture data closest to the current RGB frame timestamps, while landmarks are triangulated if sufficiently constrained.  We optimize each sequence with and without the GP depth prior to evaluate its effect.  Using the corresponding depth frames, we compare the error of the reprojected sparse landmarks where valid depths occur, as well as the dense depth map error by conditioning on the sparse landmarks.  To handle monocular scale ambiguity, we align depth maps with the optimal scale before computing the error.  A boxplot of errors between the two methods is shown in Figure \ref{fig:bundle_adj_results}.  By exploiting the correlation between observations, significantly more coherent geometry can be generated, as visualized by an example in Figure \ref{fig:qual_ba_fig}.  Note that the prior is able to achieve consistent geometry despite the low baseline and only optimizing sparse landmarks.  The shadows reveal the complete chair legs and table.  We also show a longer example with over three seconds of data fused into a TSDF in Figure \ref{fig:title_fig}.

\subsection{Monocular Dense Visual Odometry}
\label{subsec:visual_odometry}

\begin{figure*}[t]
    \includegraphics[width=\textwidth]{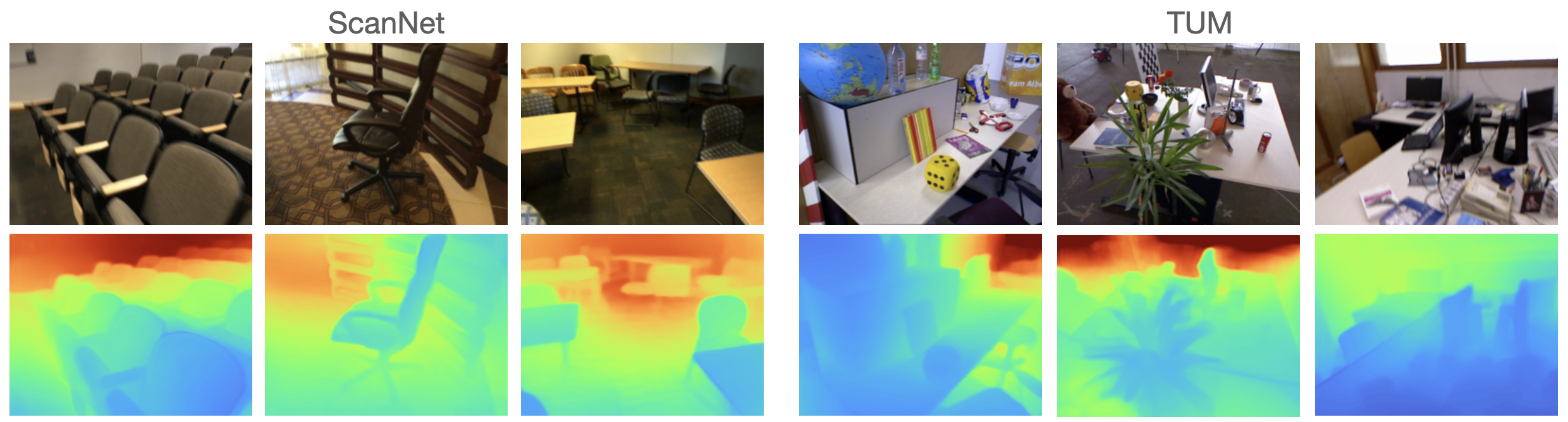}
    \caption{Example depth maps of keyframes while running the dense odometry on ScanNet validation sequences and TUM.}
    \label{fig:odom_depths} 
\end{figure*}

We propose a monocular dense visual odometry system based on the depth covariance and conditioning on per-frame latent depth points.  We perform sliding window optimization over a fixed number of keyframes, where poses, scales, and inducing depths are optimized per-frame.  To regularize the depths, we may condition on the current estimates of the inducing depths as in Section \ref{subsec:conditional}, project depths into neighboring keyframes, and apply relative photometric error constraints.  The photometric error for pixels $n$ across all directed edges $E$ is:
\begin{align}
    E = \sum_{i,j \in E} \sum_n || I_i(\mathbf{x}_n) - I_j(w(\mathbf{x}_n, \mathbf{T}_i, \mathbf{T}_j, \mathbf{y_i}, m_i) || ^2_{\sigma_r^2 I},
\end{align}
where $I$ is the image intensity, $\mathbf{x}$ are pixel coordinates, $w$ is the warping function that queries a depth map and projects the corresponding point into the neighboring image, $\mathbf{T}$ are camera poses, $\mathbf{y}$ are estimated log-depth observations, and $m$ is the mean log depth.  The corresponding factor graph with the depth priors included is shown in the right of Figure \ref{fig:ba_dvo_fg}.  To initialize a new keyframe, we first run the RGB through the network to get the covariance feature maps and sample inducing points.  We perform active sampling since this is independent of whether log-depths are known as mentioned in \ref{subsec:active_selection}, and stop sampling if a maximum variance threshold is achieved.  The log-depths $\mathbf{y}$ of a new keyframe are initialized by minimizing the least-squares cost of the prior at the sampled locations and a predictive data term over the reprojected depths $\mathbf{d}$ from the latest keyframe:
\begin{align}
    \mathbf{y}_{\text{init}} = \argmin_{\mathbf{y}}  ||\mathbf{y} - m||^2_{K} + || \log \mathbf{d} - \mathbf{f}_*||_{\text{diag}[\Sigma_{*}]}.
\end{align}
When the window size is full, the last keyframe is dropped, and scale and pose priors are added to the current estimates of the oldest keyframe still in the window.  Initialization of the system is achieved via two-frame SfM, where the first frame's depth map is optimized along with the relative pose to another frame.  Tracking is performed using coarse-to-fine photometric tracking against the last keyframe.  We also optimize affine brightness parameters for varying exposure.

For quantitative evaluation, we compare our odometry to other learning-based dense systems on Freiburg 1 sequences in the TUM dataset.  Since our method is purely a sliding-window odometry system, we compare against the comparable monocular methods evaluated in \cite{teed_droid_2021} that lack global optimization and bundle adjustment.  We omit the floor sequence which drops a significant number of frames.  The absolute trajectory errors (ATE) are shown in Table \ref{tab:tum_ate}.  Since our dense odometry does not have global scale, we align to the ground-truth trajectory's scale. 

Despite the simplicity of our system, it is first or second in 6 of 8 sequences, and achieves the best mean error.  We only use photometric error compared to the additional depth and keypoint factors in DeepFactors.  While we lack many features in our simple sliding window formulation, we believe that including additional components will further improve the results.  Qualitative examples of inferred depth maps from the odometry on ScanNet validation sequences and TUM are shown in Figure \ref{fig:odom_depths}.  Note that our representation can predict a variety of thin structures and regularize flat ones, even in the case of high-frequency texture.  

Odometry runs at an average of 7-10 Hz on $256 \times 192$ images with 8 keyframes on a single RTX 3080.  The code is not optimized and alternates between tracking and mapping in a single thread, so we believe significant speed-ups are possible given additional optimization or compute.

\setlength{\tabcolsep}{3pt}
\begin{table}[t]
    \centering
    \begin{tabular}{|c|c c c c|}
        \hline
        \multirow{2}{*}{Sequence} & TartanVO & DeepV2D & DeepFactors & Ours \\
        & \cite{wang_tartanvo_2021} & \cite{teed_deepv2d_2020} & \cite{czarnowski_deepfactors_2020} & \\
        \hline 
        360 & 0.178 & 0.243 & \underline{0.159} & \textbf{0.128} \\
        desk & \underline{0.125} & 0.166 & 0.170 & \textbf{0.056} \\
        desk2 & \underline{0.122} & 0.379 & 0.253 & \textbf{0.048} \\
        plant & 0.297 & \textbf{0.203} & 0.305 & \underline{0.261} \\
        room & 0.333 & \textbf{0.246} & 0.364 & \underline{0.257} \\
        rpy & \underline{0.049} & 0.105 & \textbf{0.043} & 0.052 \\
        teddy & \underline{0.339} & \textbf{0.316} & 0.601 & 0.475 \\
        xyz & 0.062 & 0.064 & \textbf{0.035} & \underline{0.056} \\
        \hline 
        mean & \underline{0.188} & 0.215 & 0.241 & \textbf{0.167} \\
        \hline
    \end{tabular}
    \caption{Absolute trajectory error (m) on TUM Freiburg1.  The best result is in bold, while the second best is underlined.}
    \label{tab:tum_ate}
\end{table}

%% file: sections/conclusion.tex
\section{Conclusion and Future Work}

We have proposed learning a depth covariance function, and provided selections of the depth representation, the base covariance function, and the optimization objective.  While we selected a specific nonstationary kernel function, exploration of alternatives is a valuable direction.  There is great potential to draw additional connections to the literature on Bayesian methods and GPs.  For example, geometry is low-rank and spatially local, so active selection and sparse kernels will permit scalability to higher resolutions and larger capacities.  Lastly, we demonstrated the utility of the covariance function on a selection of geometric vision tasks: depth completion, bundle adjustment, and monocular dense visual odometry.  Beyond these applications, the prior can be tightly integrated into existing geometric vision frameworks and serve as a foundation for novel formulations.

%% file: sections/supp.tex
\section{Network Architecture and Training Details}

For the convolutional neural network (CNN) component of our system, we use a UNet \cite{ronneberger_u-net_2015}.  The $256 \times 192$ RGB input is first converted into a 16-dimensional feature space, which is the input to the UNet.  We use 5 downsampling and upsampling layers.  Each downsampling layer consists of max pooling and two residual convolution layers \cite{he_deep_2016}.  Each upsampling layer bilinearly interpolates the feature map to a higher resolution, performs a convolution, and then the two residual convolutions.  We use LeakyReLU as the activation function  for all convolutions, and we also use GroupNorm \cite{wu_group_2018} with 16 groups.  

The final four upsampling layers give outputs of the covariance parameters at different resolutions.  When training, we compute the mean loss with respect to a resized depth image for each of these levels, and then scale the loss so that higher resolutions are given higher weight according to the number of pixels.  For example, the highest resolution will have four times more weight than the second highest.

For the Nystr{\"o}m approximation during training, we fix the rank to be 128.  The 128 inducing locations are sampled randomly since this is relatively efficient and was found to be more stable than active sampling.  Exploring the rank parameter and how it may trade-off expressiveness and compactness is an interesting avenue for future work. 

As mentioned previously, to handle scale, we solve for the optimal mean log-depth that minimizes the data term of the negative log marginal likelihood.  To minimize the variational free energy, we use the Adam \cite{kingma_adam_2015} optimizer with an initial learning rate of $3\mathrm{e}{-4}$.  During training, we used a batch size of 4  and performed data augmentation with random rotations, resized crops, flips, and color jitter.

\section{Active Sampling Implementation}

For active sampling of pixel locations to condition on or select as inducing points, we use the greedy variance selection described in \cite{guestrin_near-optimal_2005}. We calculate the conditional variance for all pixel locations with respect to the current samples, and select the location with the highest variance.  The conditional variance is the diagonal of the conditional covariance matrix described previously:
\begin{align}
    \Sigma_* &= K_{\text{ff}} - K_{\text{fn}} (K_{\text{nn}} + \sigma_n^2 I)^{-1} K_{\text{nf}}.
\end{align}
Since this form requires maintaining the inverse $(K_{\text{nn}} + \sigma_n^2 I)^{-1}$ which is dynamically changing, we perform $\mathcal{O}(n)$ updates to the Cholesky factorization as in \cite{ranganathan_online_2011}.  The decomposition is written as
\begin{align}
    (K_{\text{nn}} + \sigma_n^2 I)^{-1} = (L L^T)^{-1} = L^{-T} L^{-1}
\end{align}
Furthermore, we may also avoid recomputing the entire variance $\text{diag}[\Sigma_{*}]$ from scratch for each newly added point.  We may write the conditional covariance as
\begin{align}
    \Sigma_* &= K_{\text{ff}} - (L^{-1} K_{\text{nf}})^T (L^{-1} K_{\text{nf}}).
\end{align}
Since we only require $\text{diag}[\Sigma_{*}]$, we only need to add the new row of $L^{-1} K_{\text{nf}}$ for each new input point, and take the squared norm of each column when computing the variance.  We do not actually invert $L$, but instead use efficient triangular solves via forward substitution.  Thus, we only need to update $L$ and $L^{-1} K_{\text{nf}}$ each with a new row.  This avoids $\mathcal{O}(n^3)$ inversions and $\mathcal{O}(n^2)$ triangular solves for each step of the greedy selection. 

\section{Visual Odometry Photometric Factor}

As mentioned previously, given log-depths reference frame $i$, we form the photometric constraint with respect to frame $j$:
\begin{align}
    E = \sum_{i,j \in E} \sum_n || I_i(\mathbf{x}_n) - I_j(w(\mathbf{x}_n, \mathbf{T}_i, \mathbf{T}_j, \mathbf{y}_i, m_i) || ^2_{\sigma_r^2 I}
\end{align}
where $I$ is the image intensity, $\mathbf{x}$ are pixel coordinates, $w$ is the warping function that queries a depth map and projects the corresponding point into the neighboring image, $\mathbf{T}$ are camera poses, $\mathbf{y}$ are the latent log-depth observations, and $m$ is the mean log depth for a given frame.  

First, the dense log-depth map is formed via the GP conditional mean:
\begin{align}
    \mathbf{f}_* &= m_i + K_{\text{fn}} (K_{\text{nn}} + \sigma_n^2 I)^{-1} (\mathbf{y}_i - m_i).
\end{align}
The 3D points $\mathbf{P}_i$ in the reference frame can be calculated via backprojection of the vectorized pixel coordinates $\mathbf{x}_i$ via the known camera intrinsics:
\begin{align}
    \mathbf{P}_i = \pi^{-1}(\mathbf{x}_i, e^{\mathbf{f}_*}).
\end{align}
The points may then be transformed into frame $j$ and projected into the image to yield the correspondence
\begin{align}
    \mathbf{x}_j = \pi(\mathbf{T}_j^{-1} \mathbf{T}_i \mathbf{P}_i).
\end{align}
These steps describe the warping function $w$ that is used to achieve correspondence.

When the exposure times may vary, we also include affine brightness parameters $(a,b)$ for the photometric factor.  For brevity, we write the correspondences from $w$ as $\mathbf{x}_j$, so that the unwhitened residual becomes
\begin{align}
    \mathbf{r}_{i,j} = I_i(\mathbf{x}_i) + b_i - \left( \frac{e^{-a_i}}{e^{-a_j}} I_j(\mathbf{x}_j) + b_j \right).
\end{align}
The affine brightness terms are jointly optimized with the other unknowns.  To further robustify the cost against occlusion and specular surfaces, we use the Huber robust cost function instead of the non-robust least-squares cost.

%% file: main.bbl
\begin{thebibliography}{10}\itemsep=-1pt

\bibitem{bae_irondepth_2022}
Gwangbin Bae, Ignas Budvytis, and Roberto Cipolla.
\newblock {IronDepth}: Iterative refinement of single-view depth using surface
  normal and its uncertainty.
\newblock In {\em {Proceedings of the British Machine Vision Conference
  ({BMVC})}}, 2022.

\bibitem{bauer_understanding_2016}
Matthias Bauer, Mark van~der Wilk, and Carl~Edward Rasmussen.
\newblock Understanding probabilistic sparse {Gaussian} process approximations.
\newblock In {\em {Neural Information Processing Systems ({NeurIPS})}}, 2016.

\bibitem{bloesch_codeslam_2018}
Michael Bloesch, Jan Czarnowski, Ronald Clark, Stefan Leutenegger, and
  Andrew~J. Davison.
\newblock {CodeSLAM} - learning a compact, optimisable representation for dense
  visual {SLAM}.
\newblock In {\em {Proceedings of the {IEEE} Conference on Computer Vision and
  Pattern Recognition ({CVPR})}}, 2018.

\bibitem{bloesch_learning_2019}
Michael Bloesch, Tristan Laidlow, Ronald Clark, Stefan Leutenegger, and Andrew
  Davison.
\newblock Learning meshes for dense visual {SLAM}.
\newblock In {\em {Proceedings of the International Conference on Computer
  Vision ({ICCV})}}, 2019.

\bibitem{chen_ak_2022}
Weizhe Chen, Roni Khardon, and Lantao Liu.
\newblock {AK}: Attentive kernel for information gathering.
\newblock In {\em {Proceedings of Robotics: Science and Systems ({RSS})}},
  2022.

\bibitem{cheng_depth_2018}
Xinjing Cheng, Peng Wang, and Ruigang Yang.
\newblock Depth estimation via affinity learned with convolutional spatial
  propagation network.
\newblock In {\em {Proceedings of the European Conference on Computer Vision
  ({ECCV})}}, 2018.

\bibitem{civera_inverse_2008}
Javier Civera, Andrew~J. Davison, and J.~M.~MartÍnez Montiel.
\newblock Inverse depth parametrization for monocular {SLAM}.
\newblock {\em {{IEEE} Transactions on Robotics ({T-RO})}}, 2008.

\bibitem{conti_sparsity_2023}
Andrea Conti, Matteo Poggi, and Stefano Mattoccia.
\newblock Sparsity agnostic depth completion.
\newblock In {\em {Proceedings of the {IEEE/CVF} Winter Conference on
  Applications of Computer Vision ({WACV})}}, 2023.

\bibitem{czarnowski_deepfactors_2020}
Jan Czarnowski, Tristan Laidlow, Ronald Clark, and Andrew~J. Davison.
\newblock {DeepFactors}: Real-time probabilistic dense monocular {SLAM}.
\newblock {\em {{IEEE} Robotics and Automation Letters ({RA-L})}}, 2020.

\bibitem{dai_scannet_2017}
Angela Dai, Angel~X. Chang, Manolis Savva, Maciej Halber, Thomas Funkhouser,
  and Matthias Nie{\ss}ner.
\newblock {ScanNet}: Richly-annotated {3D} reconstructions of indoor scenes.
\newblock In {\em {Proceedings of the {IEEE} Conference on Computer Vision and
  Pattern Recognition ({CVPR})}}, 2017.

\bibitem{dorta_structured_2018}
Garoe Dorta, Sara Vicente, Lourdes Agapito, Neill D.~F. Campbell, and Ivor
  Simpson.
\newblock Structured uncertainty prediction networks.
\newblock In {\em {Proceedings of the {IEEE} Conference on Computer Vision and
  Pattern Recognition ({CVPR})}}, 2018.

\bibitem{eigen_depth_2014}
David Eigen, Christian Puhrsch, and Rob Fergus.
\newblock Depth map prediction from a single image using a multi-scale deep
  network.
\newblock In {\em {Neural Information Processing Systems ({NeurIPS})}}, 2014.

\bibitem{engel_lsd-slam_2014}
Jakob Engel, Thomas Schöps, and Daniel Cremers.
\newblock {LSD}-{SLAM}: Large-scale direct monocular {SLAM}.
\newblock In {\em {Proceedings of the European Conference on Computer Vision
  ({ECCV})}}, 2014.

\bibitem{dellaert_gtsam_2022}
Frank~Dellaert et al.
\newblock borglab/gtsam, May 2022.

\bibitem{godard_digging_2019}
Clement Godard, Oisin Mac~Aodha, Michael Firman, and Gabriel~J. Brostow.
\newblock Digging into self-supervised monocular depth estimation.
\newblock In {\em {Proceedings of the International Conference on Computer
  Vision ({ICCV})}}, 2019.

\bibitem{graham_ridgesfm_2020}
Benjamin Graham and David Novotny.
\newblock {RidgeSfM}: Structure from motion via robust pairwise matching under
  depth uncertainty.
\newblock In {\em {Proceedings of the International Conference on 3D Vision
  ({3DV})}}, 2020.

\bibitem{guestrin_near-optimal_2005}
Carlos Guestrin, Andreas Krause, and Ajit~Paul Singh.
\newblock Near-optimal sensor placements in {Gaussian} processes.
\newblock In {\em {Proceedings of the International Conference on Machine
  Learning ({ICML})}}, 2005.

\bibitem{he_deep_2016}
K. He, X. Zhang, S. Ren, and J. Sun.
\newblock Deep residual learning for image recognition.
\newblock In {\em {Proceedings of the {IEEE} Conference on Computer Vision and
  Pattern Recognition ({CVPR})}}, 2016.

\bibitem{jebara_probability_2004}
Tony Jebara, Risi Kondor, and Andrew Howard.
\newblock Probability product kernels.
\newblock {\em {{The Journal of Machine Learning Research}}}, 2004.

\bibitem{kendall_what_2017}
Alex Kendall and Yarin Gal.
\newblock What uncertainties do we need in {Bayesian} deep learning for
  computer vision?
\newblock In {\em {Neural Information Processing Systems ({NeurIPS})}}, 2017.

\bibitem{kingma_adam_2015}
Diederik~P. Kingma and Jimmy Ba.
\newblock Adam: {A} method for stochastic optimization.
\newblock In {\em {Proceedings of the International Conference on Learning
  Representations ({ICLR})}}, 2015.

\bibitem{kingma_autoencoding_2014}
Diederik~P. Kingma and Max Welling.
\newblock Auto-encoding variational {Bayes}.
\newblock In {\em {Proceedings of the International Conference on Learning
  Representations ({ICLR})}}, 2014.

\bibitem{kuleshov_accurate_2018}
Volodymyr Kuleshov, Nathan Fenner, and Stefano Ermon.
\newblock Accurate uncertainties for deep learning using calibrated regression.
\newblock In {\em {Proceedings of the International Conference on Machine
  Learning ({ICML})}}, 2018.

\bibitem{lang_adaptive_2007}
Tobias Lang, Christian Plagemann, and Wolfram Burgard.
\newblock Adaptive non-stationary kernel regression for terrain modeling.
\newblock In {\em {Proceedings of Robotics: Science and Systems ({RSS})}},
  2007.

\bibitem{liu_learning_2021}
Lina Liu, Yiyi Liao, Yue Wang, Andreas Geiger, and Yong Liu.
\newblock Learning steering kernels for guided depth completion.
\newblock {\em {{IEEE} Transactions on Image Processing}}, 2021.

\bibitem{lucas_iterative_1981}
Bruce~D. Lucas and Takeo Kanade.
\newblock An iterative image registration technique with an application to
  stereo vision.
\newblock In {\em {Proceedings of the International Joint Conference on
  Artificial Intelligence ({IJCAI})}}, 1981.

\bibitem{ma_sparse_2018}
Fangchang Ma and Sertac Karaman.
\newblock Sparse-to-dense: Depth prediction from sparse depth samples and a
  single image.
\newblock In {\em {Proceedings of the {IEEE} International Conference on
  Robotics and Automation ({ICRA})}}, 2018.

\bibitem{mur-artal_orb-slam2_2017}
Raúl Mur-Artal and Juan~D. Tardós.
\newblock {ORB}-{SLAM2}: An open-source {SLAM} system for monocular, stereo,
  and {RGB}-{D} cameras.
\newblock {\em {{IEEE} Transactions on Robotics ({T-RO})}}, 2017.

\bibitem{newcombe_dtam_2011}
Richard.~A. Newcombe, Steven~J. Lovegrove, and Andrew~J. Davison.
\newblock {DTAM}: Dense tracking and mapping in real-time.
\newblock In {\em {Proceedings of the International Conference on Computer
  Vision ({ICCV})}}, 2011.

\bibitem{ober_promises_2021}
Sebastian~W. Ober, Carl~E. Rasmussen, and Mark van~der Wilk.
\newblock The promises and pitfalls of deep kernel learning.
\newblock In {\em {Conference on Uncertainty in Artificial Intelligence}},
  2021.

\bibitem{paciorek_nonstationary_2003}
Christopher Paciorek and Mark Schervish.
\newblock Nonstationary covariance functions for {Gaussian} process regression.
\newblock In {\em {Neural Information Processing Systems ({NeurIPS})}}, 2003.

\bibitem{park_nonlocal_2020}
Jinsun Park, Kyungdon Joo, Zhe Hu, Chi-Kuei Liu, and In So~Kweon.
\newblock Non-local spatial propagation network for depth completion.
\newblock In {\em {Proceedings of the European Conference on Computer Vision
  ({ECCV})}}, 2020.

\bibitem{patacchiola_bayesian_2020}
Massimiliano Patacchiola, Jack Turner, Elliot~J. Crowley, Michael O'~Boyle, and
  Amos~J Storkey.
\newblock {Bayesian} meta-learning for the few-shot setting via deep kernels.
\newblock In {\em {Neural Information Processing Systems ({NeurIPS})}}, 2020.

\bibitem{ranftl_towards_2022}
Rene Ranftl, Katrin Lasinger, David Hafner, Konrad Schindler, and Vladlen
  Koltun.
\newblock Towards robust monocular depth estimation: Mixing datasets for
  zero-shot cross-dataset transfer.
\newblock {\em {{IEEE} Transactions on Pattern Analysis and Machine
  Intelligence ({PAMI})}}, 2022.

\bibitem{ranganathan_online_2011}
Ananth Ranganathan, Ming-Hsuan Yang, and Jeffrey Ho.
\newblock Online sparse {Gaussian} process regression and its applications.
\newblock {\em {{IEEE} Transactions on Image Processing}}, 2011.

\bibitem{rasmussen_gaussian_2005}
Carl~Edward Rasmussen and Christopher K.~I. Williams.
\newblock {\em {Gaussian} {Processes} for Machine Learning}.
\newblock MIT Press, 2005.

\bibitem{rendu_normal_1979}
Jean-Michel~M. Rendu.
\newblock Normal and lognormal estimation.
\newblock {\em Journal of the International Association for Mathematical
  Geology}, 1979.

\bibitem{ronneberger_u-net_2015}
Olaf Ronneberger, Philipp Fischer, and Thomas Brox.
\newblock U-{Net}: Convolutional networks for biomedical image segmentation.
\newblock In {\em {Proceedings of the International Conference on Medical Image
  Computing and Computer Assisted Intervention ({MICCAI})}}, 2015.

\bibitem{schonberger_structure--motion_2016}
Johannes~L. Schonberger and Jan-Michael Frahm.
\newblock Structure-from-motion revisited.
\newblock In {\em {Proceedings of the {IEEE} Conference on Computer Vision and
  Pattern Recognition ({CVPR})}}, 2016.

\bibitem{shi_good_1994}
J. Shi and C. Tomasi.
\newblock Good features to track.
\newblock In {\em {Proceedings of the {IEEE} Conference on Computer Vision and
  Pattern Recognition ({CVPR})}}, 1994.

\bibitem{silberman_indoor_2012}
Nathan Silberman, Derek Hoiem, Pushmeet Kohli, and Rob Fergus.
\newblock Indoor segmentation and support inference from {RGBD} images.
\newblock In {\em {Proceedings of the European Conference on Computer Vision
  ({ECCV})}}, 2012.

\bibitem{simpson_learning_2022}
Ivor J.~A. Simpson, Sara Vicente, and Neill D.~F. Campbell.
\newblock Learning structured {Gaussians} to approximate deep ensembles.
\newblock In {\em {Proceedings of the {IEEE} Conference on Computer Vision and
  Pattern Recognition ({CVPR})}}, 2022.

\bibitem{sturm_benchmark_2012}
Jürgen Sturm, Nikolas Engelhard, Felix Endres, Wolfram Burgard, and Daniel
  Cremers.
\newblock A benchmark for the evaluation of {RGB-D} {SLAM} systems.
\newblock In {\em {Proceedings of the {IEEE/RSJ} Conference on Intelligent
  Robots and Systems ({IROS})}}, 2012.

\bibitem{tang_ba-net_2019}
Chengzhou Tang and Ping Tan.
\newblock {BA}-{Net}: Dense bundle adjustment networks.
\newblock In {\em {Proceedings of the International Conference on Learning
  Representations ({ICLR})}}, 2019.

\bibitem{teed_deepv2d_2020}
Zachary Teed and Jia Deng.
\newblock {DeepV2D}: Video to depth with differentiable structure from motion.
\newblock In {\em {Proceedings of the International Conference on Learning
  Representations ({ICLR})}}, 2020.

\bibitem{teed_droid_2021}
Zachary Teed and Jia Deng.
\newblock {DROID-SLAM}: Deep visual {SLAM} for monocular, stereo, and {RGB-D}
  cameras.
\newblock In {\em {Neural Information Processing Systems ({NeurIPS})}}, 2021.

\bibitem{titsias_variational_2009}
Michalis Titsias.
\newblock Variational learning of inducing variables in sparse {Gaussian}
  processes.
\newblock In {\em {International Conference on Artificial Intelligence and
  Statistics ({AISTATS})}}, 2009.

\bibitem{wang_tartanvo_2021}
Wenshan Wang, Yaoyu Hu, and Sebastian Scherer.
\newblock {TartanVO}: A generalizable learning-based {VO}.
\newblock In {\em {Conference on Robot Learning ({CoRL})}}, 2021.

\bibitem{wilson_deep_2016}
Andrew~Gordon Wilson, Zhiting Hu, Ruslan Salakhutdinov, and Eric~P. Xing.
\newblock Deep kernel learning.
\newblock In {\em {International Conference on Artificial Intelligence and
  Statistics ({AISTATS})}}, 2016.

\bibitem{wu_group_2018}
Yuxin Wu and Kaiming He.
\newblock Group normalization.
\newblock In {\em {Proceedings of the European Conference on Computer Vision
  ({ECCV})}}, September 2018.

\bibitem{yan_rignet_2022}
Zhiqiang Yan, Kun Wang, Xiang Li, Zhenyu Zhang, Jun Li, and Jian Yang.
\newblock {RigNet}: Repetitive image guided network for depth completion.
\newblock In {\em {Proceedings of the European Conference on Computer Vision
  ({ECCV})}}, 2022.

\bibitem{yoon_balanced_2020}
Sungho Yoon and Ayoung Kim.
\newblock Balanced depth completion between dense depth inference and sparse
  range measurements via {KISS}-{GP}.
\newblock In {\em {Proceedings of the {IEEE/RSJ} Conference on Intelligent
  Robots and Systems ({IROS})}}, 2020.

\end{thebibliography}
